\documentclass{article}
\usepackage{spconf,amsmath,graphicx,adjustbox,svg,subcaption,booktabs}
\usepackage[noadjust]{cite}

\title{Perception-Distortion Trade-off in the SR Space Spanned by Flow Models}
\name{Cansu Korkmaz$^a$, A.Murat Tekalp$^a$, Zafer Doğan$^a$, Erkut Erdem$^b$, Aykut Erdem$^a$\thanks{This work was supported in part by an AI Fellowship to C. Korkmaz provided by the KUIS AI Center. This work was supported in part by TUBITAK 2247-A Award No.~120C156, TUBITAK 2232 Award No.~118C337, and KUIS AI Center funded by Turkish Is Bank. AMT acknowledges support from Turkish Academy of Sciences (TUBA), and AE~acknowledges BAGEP Award of the Science Academy.}}
\address{$^a$ College of Engineering and KUIS AI Center, 
Koç University, Istanbul, Turkey\\$^b$ Department of Computer Engineering, Hacettepe University, Ankara, Turkey}
\begin{document}
%
\maketitle

\begin{abstract}
Flow-based generative super-resolution (SR) models learn to produce a diverse set of feasible SR solutions, called the SR space. Diversity of SR solutions increases with the temperature ($\tau$) of latent variables, which introduces random variations of texture among sample solutions, resulting in visual artifacts and low fidelity. In this paper, we present a simple but effective image ensembling/fusion approach to obtain a single SR image eliminating random artifacts and improving fidelity without significantly compromising perceptual quality. We achieve this by benefiting from a diverse set of feasible photo-realistic solutions in the SR space spanned by flow models. We propose different image ensembling and fusion strategies which offer multiple paths to move sample solutions in the SR space to more desired destinations in the perception-distortion plane in a controllable manner depending on the fidelity vs.\ perceptual quality requirements of the task at hand. Experimental results demonstrate that our image ensembling/fusion strategy achieves more promising perception-distortion trade-off compared to sample SR images produced by flow models and adversarially trained models in terms of both quantitative metrics and visual quality. 
\end{abstract}
\begin{keywords}
normalizing flows, super-resolution, image ensembles, image fusion, perception-distortion trade-off
\end{keywords}

\section{Introduction}
\label{sec:intro}
Deep-learning based super-resolution (SR) methods have produced astonishing results~\cite{dong2015image, kim2016accurate, ledig2017photorealistic, EDSR2017, srcnn,  RCAN2018, wang2018esrgan, Wei_2021_CVPR}. Given LR-HR paired training data, early deep SR methods~\cite{dong2015image, kim2016accurate, ledig2017photorealistic, EDSR2017, srcnn,  RCAN2018} considered SR as a regression problem and produced a single deterministic output. Recently, several flow-based methods \cite{lugmayr2020srflow, jo2021srflowda} and challenges \cite{Lugmayr_2021_CVPR} recognized the ill-posed nature of the SR problem and proposed learning a one-to-many stochastic generative mapping to produce a diverse set of plausible SR images, called the SR space. These formulations consider photo-realism of solutions, their consistency with the LR image, and how well they span the SR space. 
Even though generative models can generate a diverse set of possible SR solutions, they introduce a new problem: selecting a single solution when the goal is to extract critical information from the SR image, e.g., whether a digit is 3 or 8. In such information-centric applications, generating multiple feasible solutions may not lead to a decisive result and a single high fidelity solution is desirable. While in some other applications, obtaining the~best (single) photo-realistic natural image may be more desirable.

\begin{figure}[!t]
\centering
\includegraphics[width=0.5\textwidth]{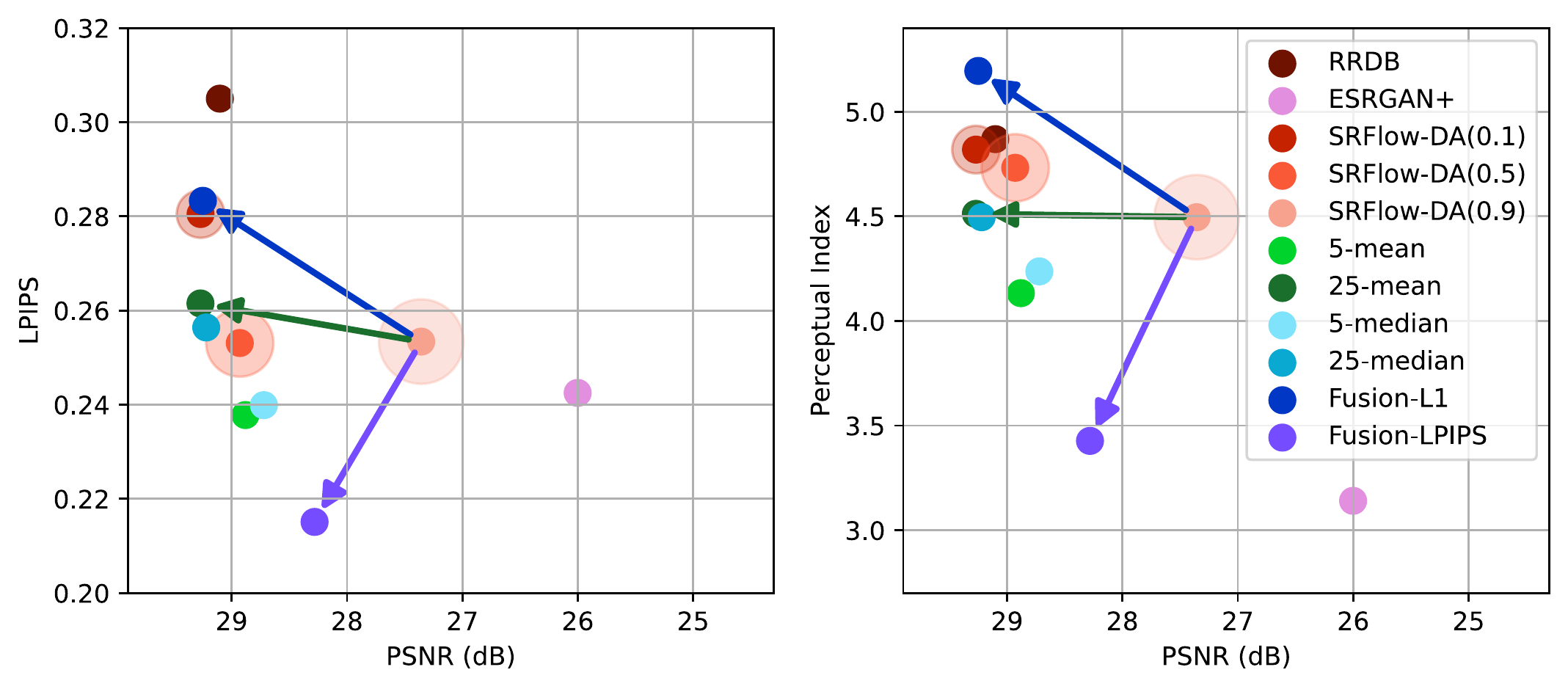} 
\vspace{-22pt}
\caption{\textbf{Perception-distortion trade-off for different SR architectures.} RRDB provides high-fidelity outputs that lack fine details. ESRGAN+ and SRFlow-DA provide sharpness by hallucinating high-frequency details, thus, have poor fidelity. Shadows around flow results indicate sample image variance induced by the random seeds. The proposed image-fusion approach offers controllable trade-off between fidelity and perception. The arrows show different trade-off paths.}
\label{fig:perception_distortion2} \vspace{-8pt}
\end{figure}

\begin{figure*}[ht]
\centering
\includegraphics[width=\linewidth]{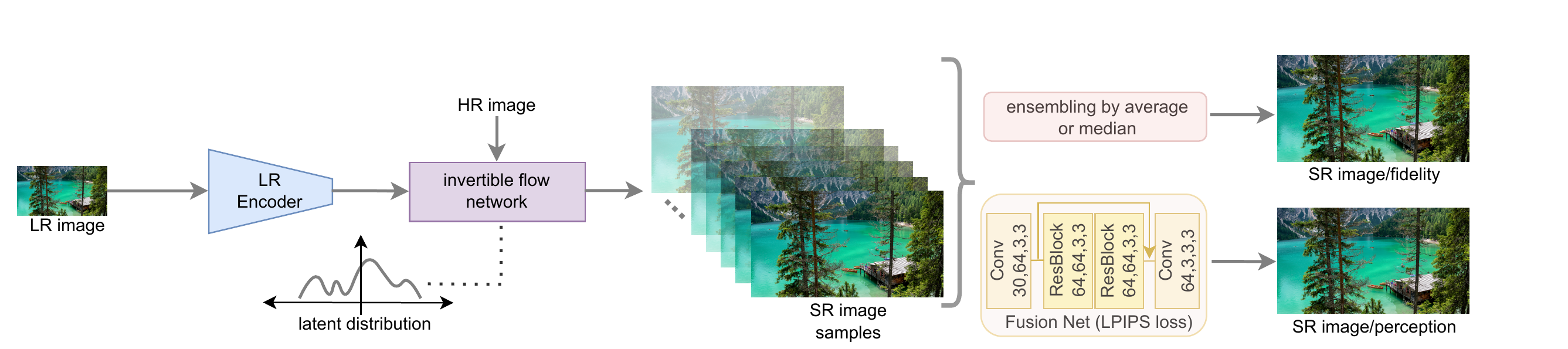} \vspace{-18pt}
\caption{\textbf{Block diagram of the proposed approach for perception-distortion trade-off in the SR space}. The trade-off path can be steered either towards higher fidelity by ensembling samples generated by a flow model using average or median operations or towards better perceptual quality using a simple fusion ConvNet trained by LPIPS loss. 
} \vspace{-10pt}
\label{fig:fusion_arch}
\end{figure*}

To this effect, we propose first generating finitely many samples in the SR space spanned by flow models, and then merging or fusing them as a means of obtaining the best perception-distortion trade-off for the application at hand, instead of searching for a single sample realization among infinitely many feasible solutions. Interestingly, our perception-distortion trade-off results summarized in Fig.~\ref{fig:perception_distortion2} show that fusing multiple images in the SR space via simple integration techniques leads to better PSNR than that of the best regressive convolutional network (ConvNet) model, e.g., RRDB~\cite{wang2018esrgan} and any feasible SRFlow sample without a significant compromise in perceptual quality of solutions with the best perceptual quality. Furthermore, we achieve these results by simple image integration techniques in the SR space instead of retraining the entire generative model as would be necessary in adversarially trained models, e.g., ESRGAN+~\cite{esrganplus}. 

In summary, our main contributions are:\\
1. We propose a novel approach for perception-distortion trade-off in the SR space spanned by flow models by means of ensembling or fusing multiple sample SR solutions.\\
2. We show that ensembling by simple averaging or median operations over samples in the SR space lead to higher fidelity results without significant degradation of perceptual quality. \\
3. We propose a simple ConvNet consisting of 2-4 residual blocks that learns fusing samples in the SR space using LPIPS loss to favor the~perception dimension for natural images. 
\vspace{-10pt}
\section{Related Work}
\label{sec:related_works}
\vspace{-6pt}
\noindent \textbf{Regressive Inference.} Several ConvNet based single image SR models, including SRCNN \cite{dong2015image}, SRResNet \cite{ledig2017photorealistic}, EDSR \cite{EDSR2017} and RCAN \cite{RCAN2018}, focus on learning a regressive LR-to-HR mapping, which is optimized for L1 or L2 reconstruction loss. 
These approaches produce results that tend towards the mean of a set of plausible SR images, which is high fidelity in terms of PSNR but appears overly smoothed and blurry. \vspace{3pt}

\begin{figure*}[!t]
\centering
\begin{subfigure}{0.16\textwidth}
    \begin{subfigure}{\textwidth}
        \includegraphics[width=\textwidth]{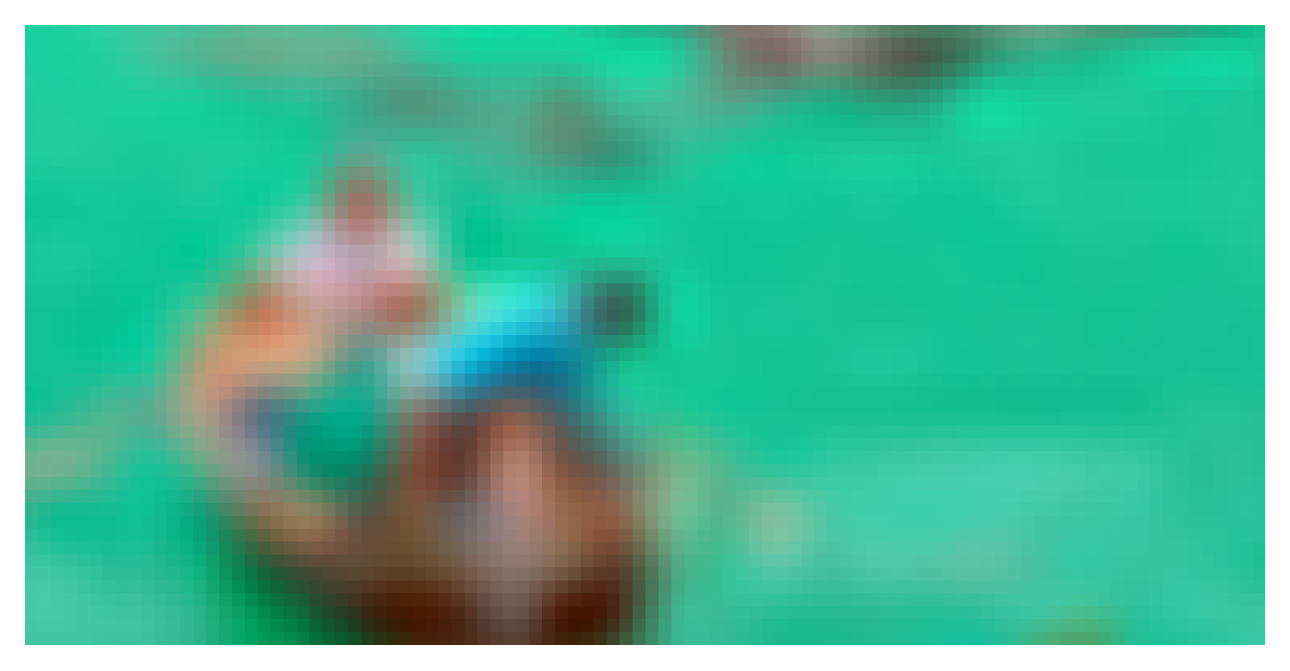}
    \end{subfigure}
    \begin{subfigure}{\textwidth}
        \includegraphics[width=\textwidth]{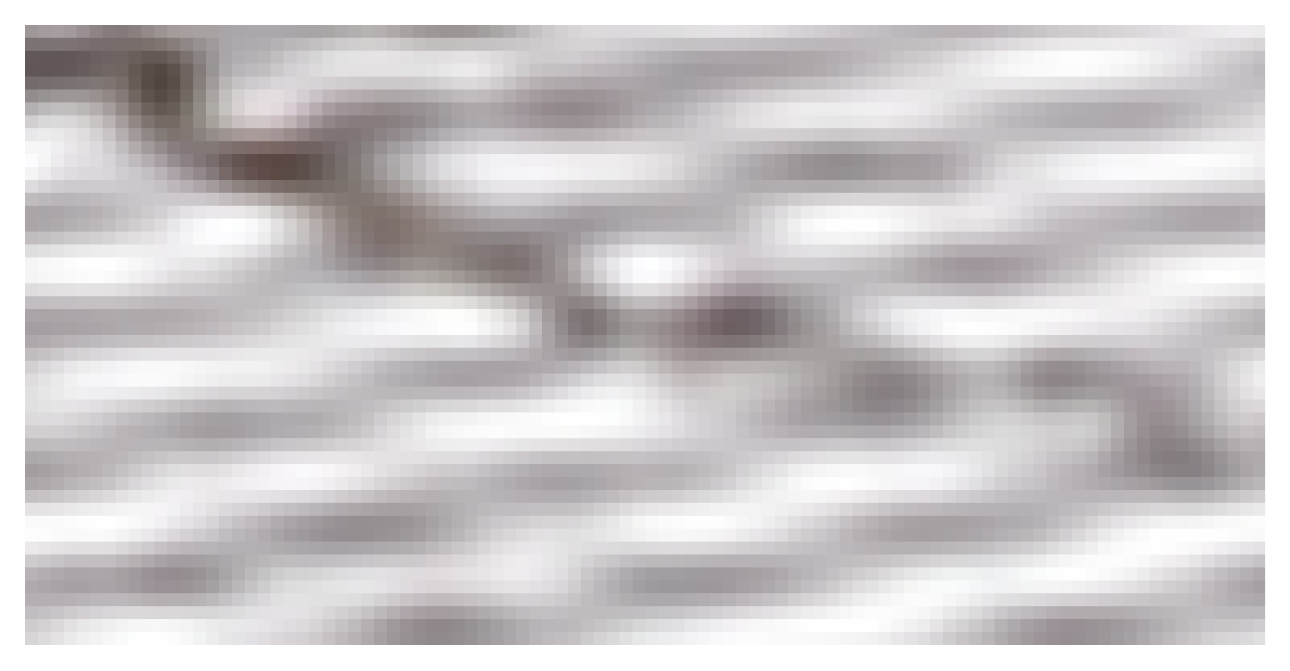}
    \end{subfigure}
        \begin{subfigure}{\textwidth}
        \includegraphics[width=\textwidth]{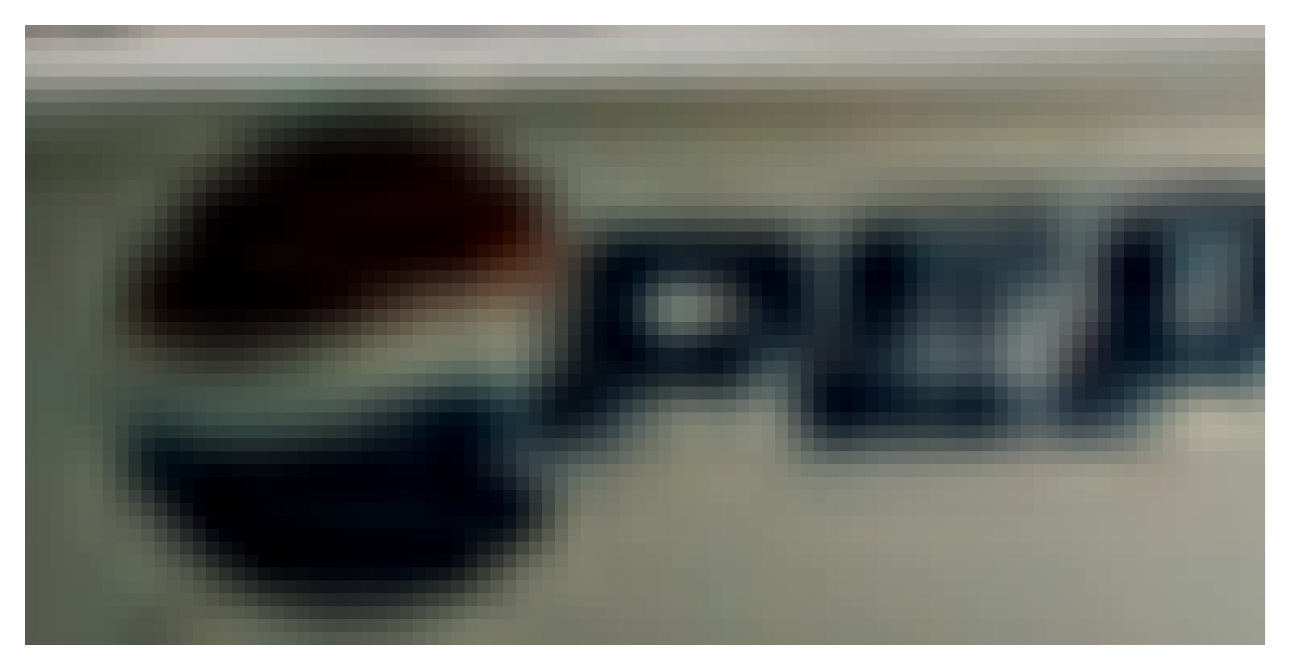}
    \end{subfigure} \vspace{-5pt}
\caption{LR - Bicubic}
\end{subfigure}
\begin{subfigure}{0.16\textwidth}
    \begin{subfigure}{\textwidth}
        \includegraphics[width=\textwidth]{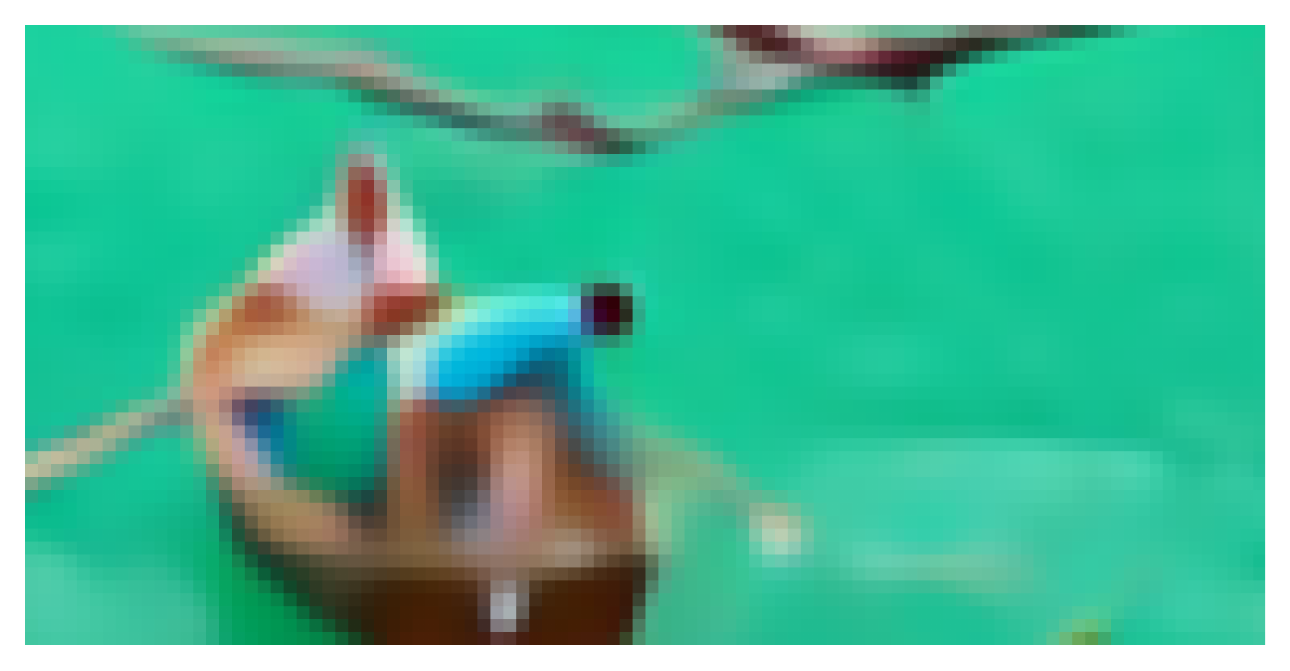}
    \end{subfigure}
    \begin{subfigure}{\textwidth}
        \includegraphics[width=\textwidth]{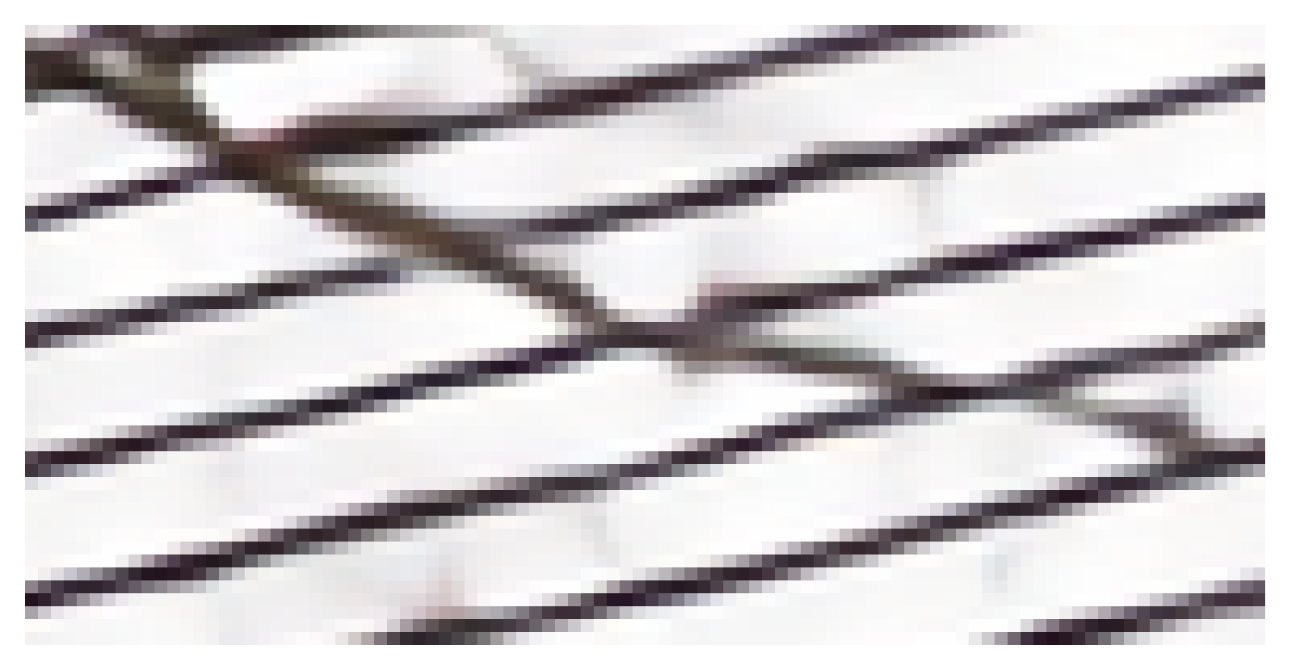}
    \end{subfigure}
        \begin{subfigure}{\textwidth}
        \includegraphics[width=\textwidth]{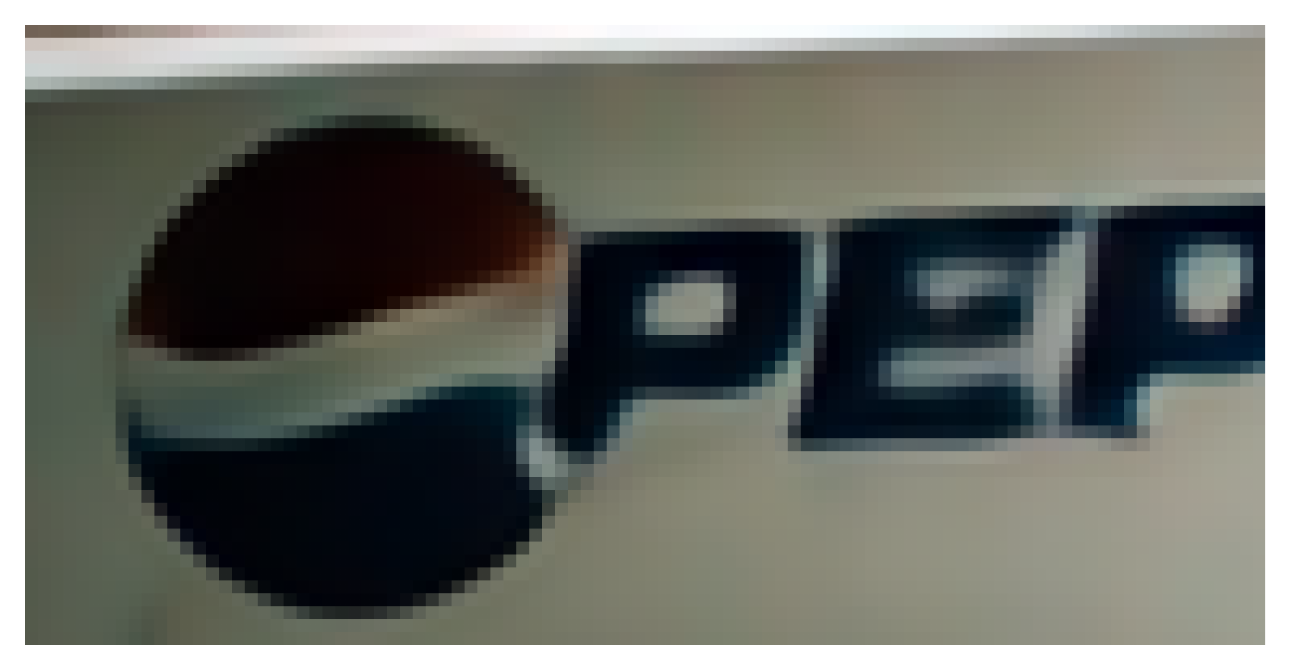}
    \end{subfigure} \vspace{-5pt}
\caption{RRDB \cite{wang2018esrgan}}
\end{subfigure}
\begin{subfigure}{0.16\textwidth}
    \begin{subfigure}{\textwidth}
        \includegraphics[width=\textwidth]{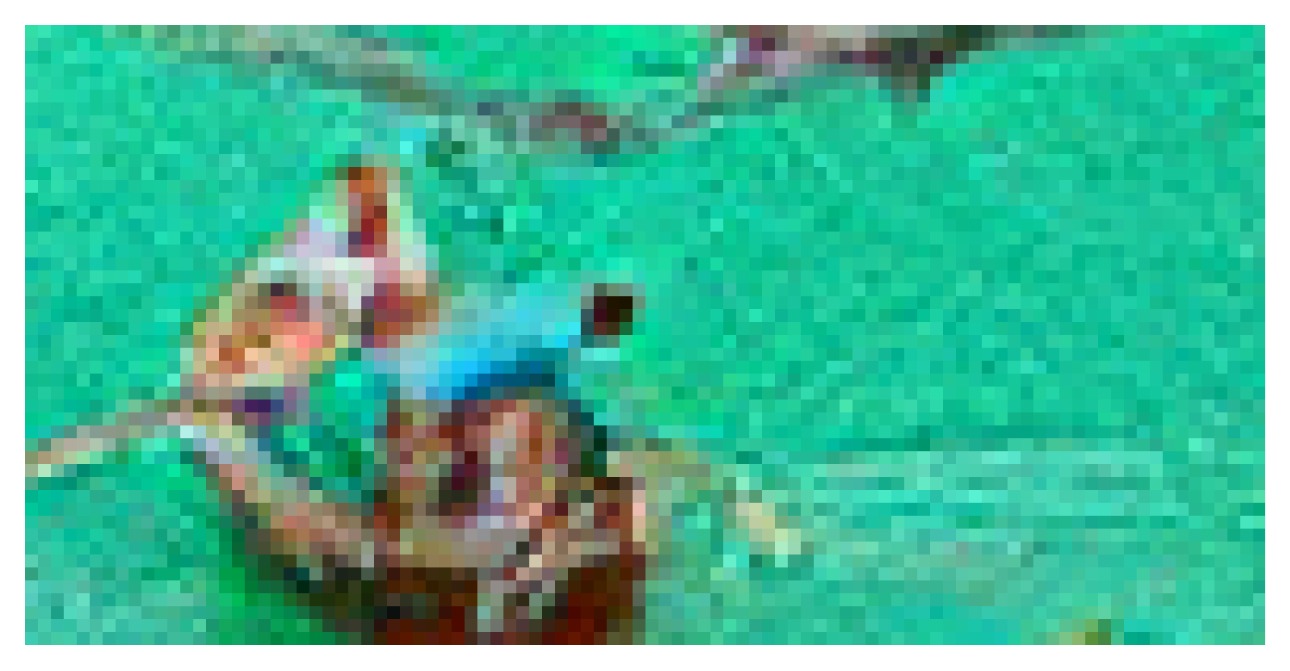}
    \end{subfigure}
    \begin{subfigure}{\textwidth}
        \includegraphics[width=\textwidth]{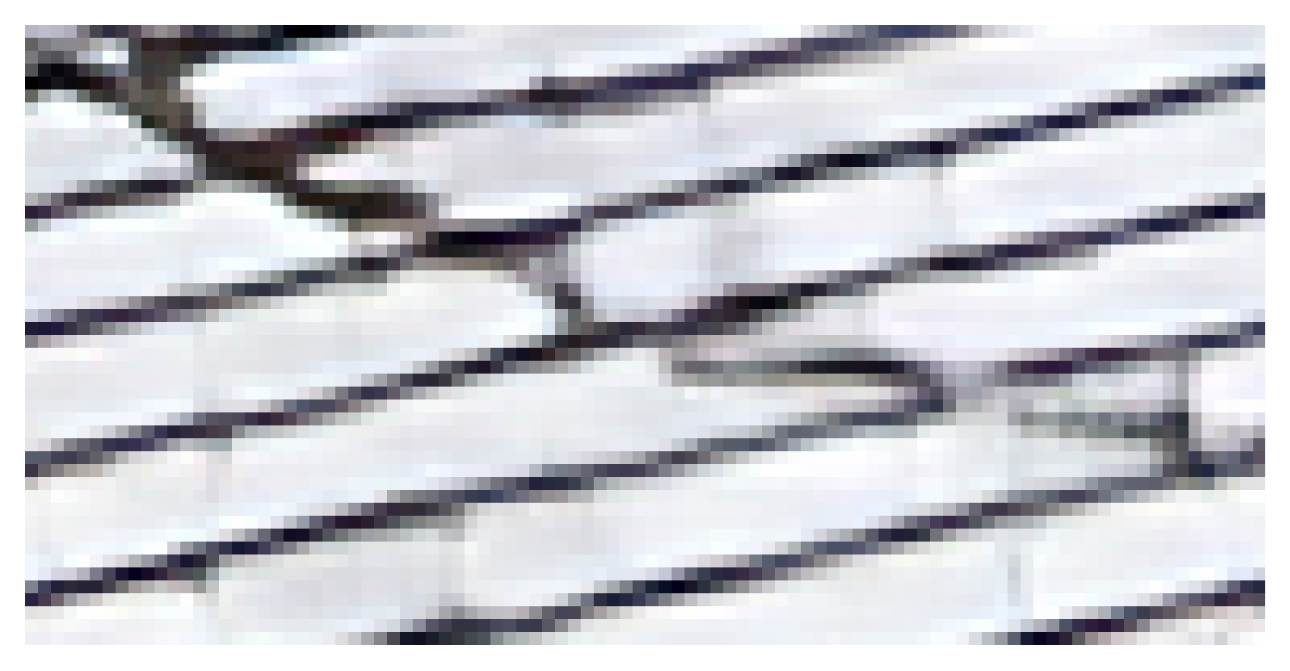}
    \end{subfigure}
        \begin{subfigure}{\textwidth}
        \includegraphics[width=\textwidth]{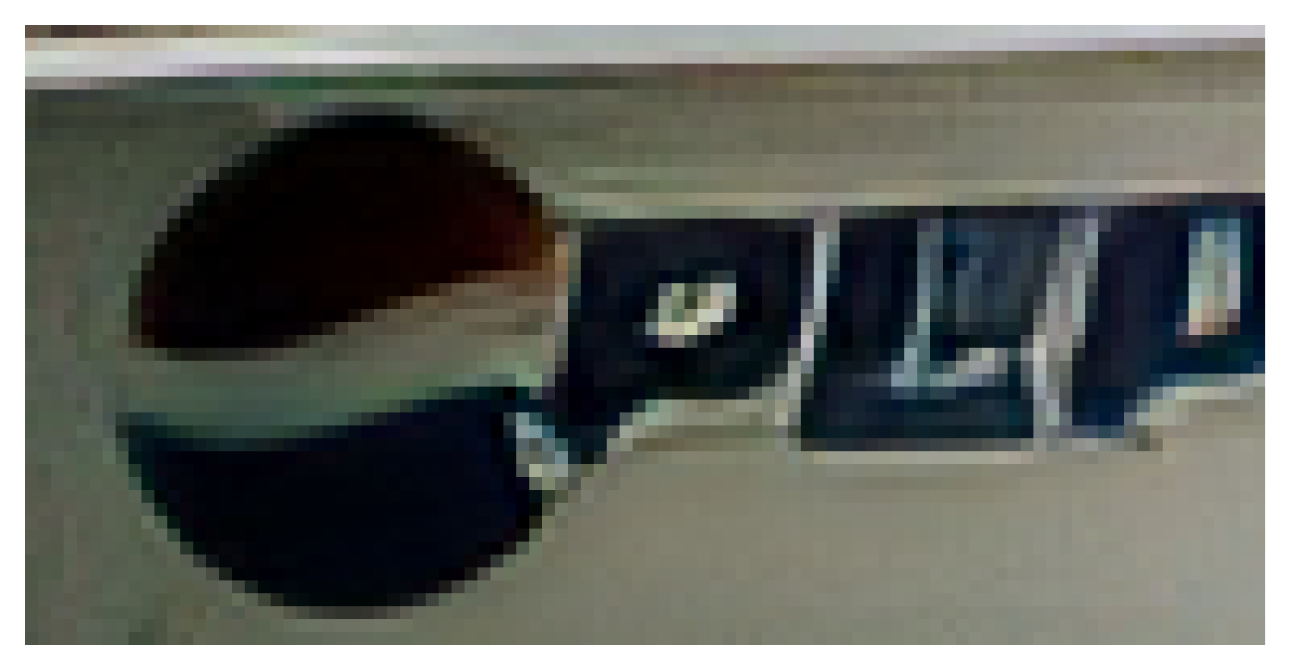}
    \end{subfigure} \vspace{-5pt}
\caption{ESRGAN+ \cite{esrganplus}}
\end{subfigure}
\begin{subfigure}{0.16\textwidth}
    \begin{subfigure}{\textwidth}
        \includegraphics[width=\textwidth]{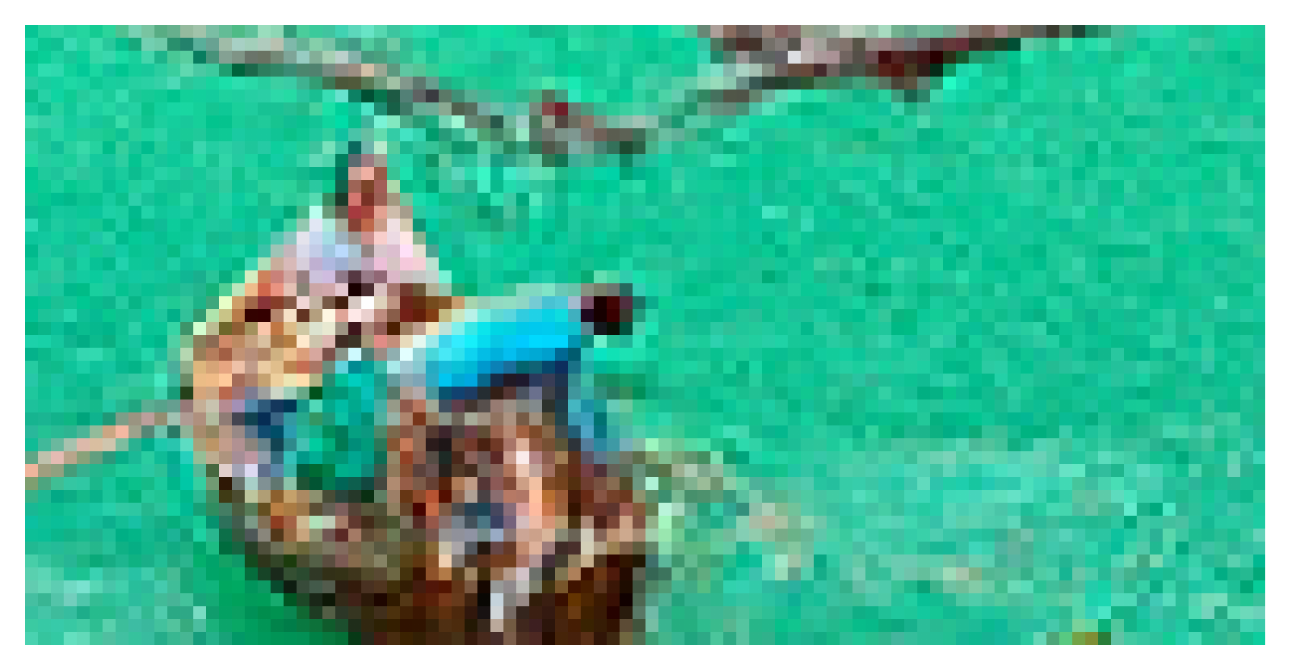}
    \end{subfigure}
    \begin{subfigure}{\textwidth}
        \includegraphics[width=\textwidth]{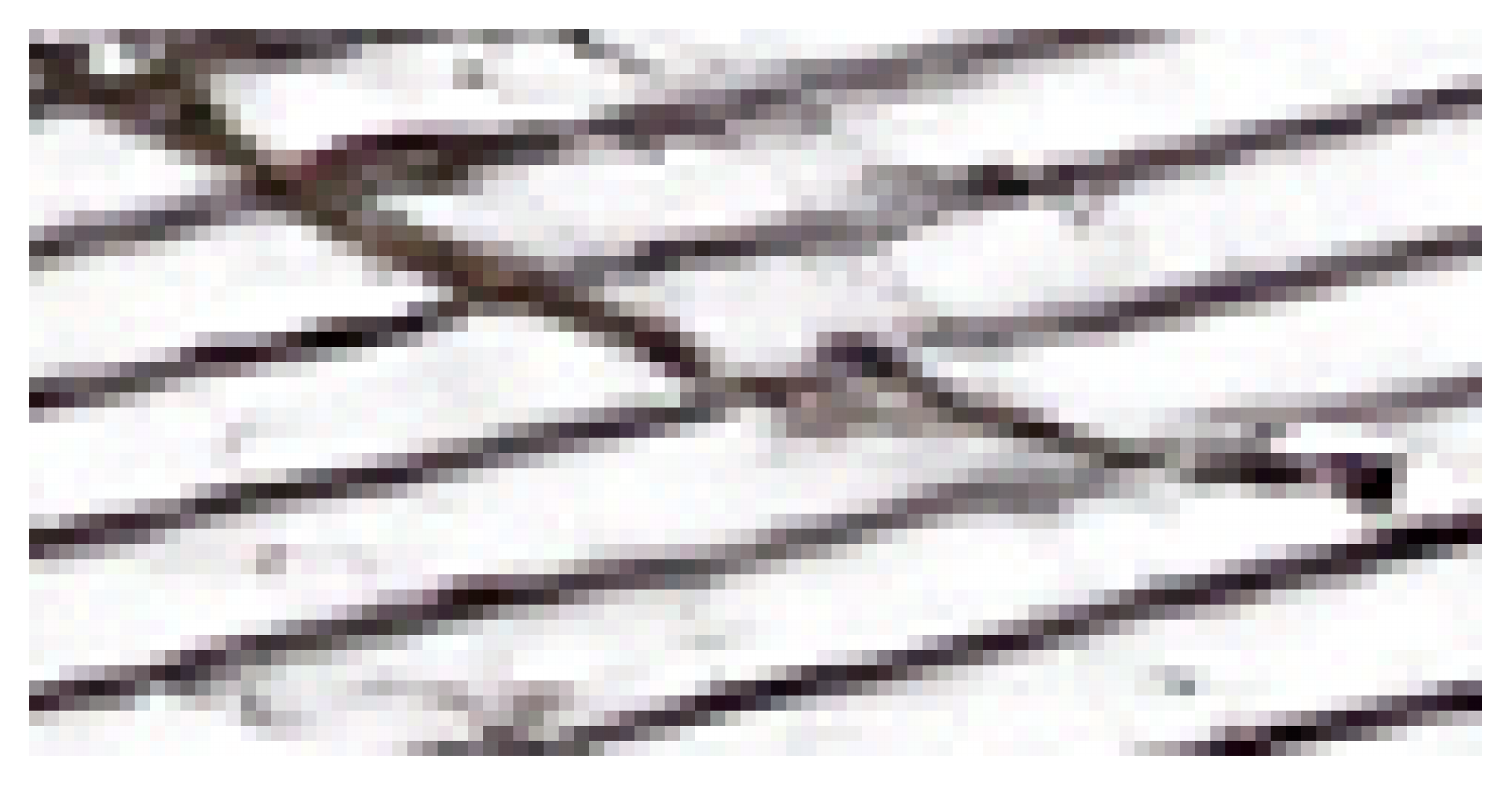}
    \end{subfigure}
    \begin{subfigure}{\textwidth}
        \includegraphics[width=\textwidth]{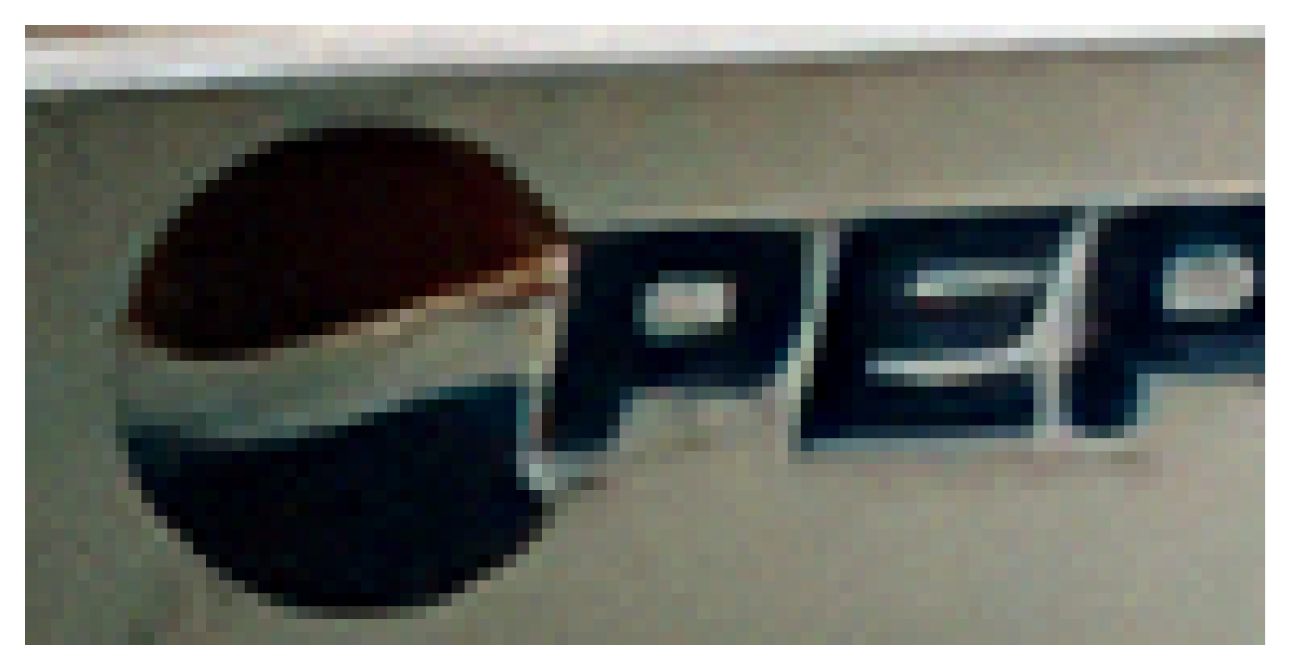}
    \end{subfigure} \vspace{-5pt}
\caption{SRFlow-DA \cite{jo2021srflowda}}
\end{subfigure}
\begin{subfigure}{0.16\textwidth}
    \begin{subfigure}{\textwidth}
        \includegraphics[width=\textwidth]{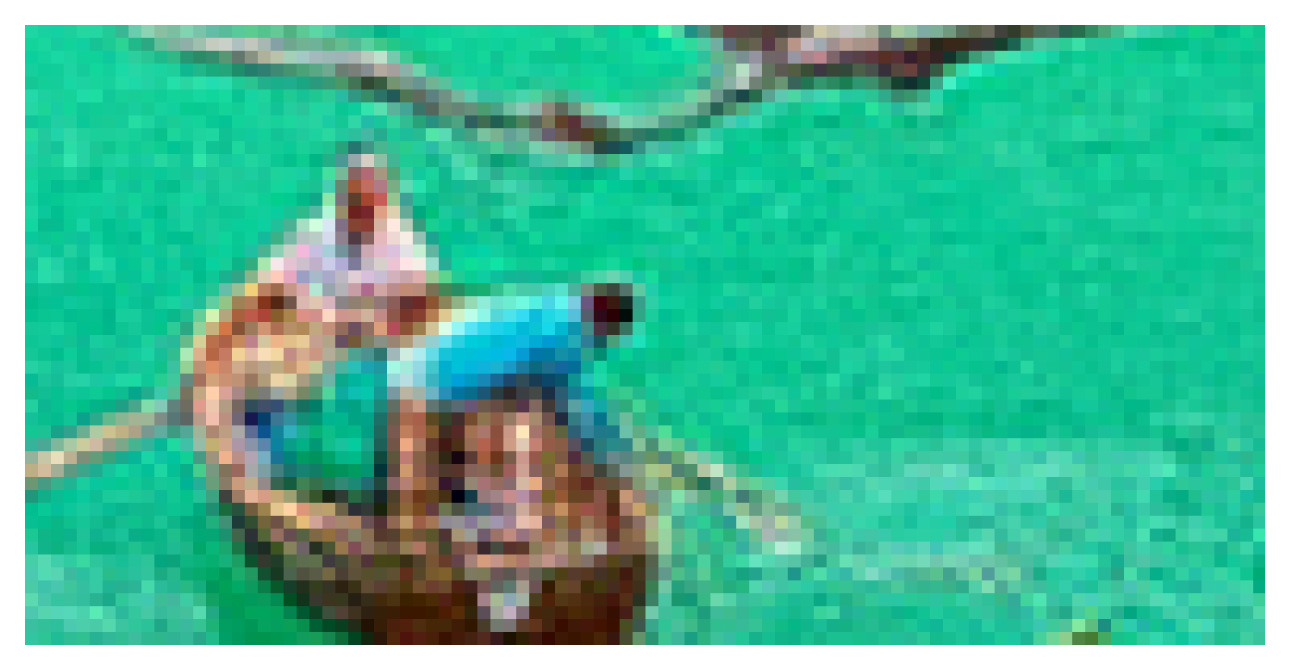}
    \end{subfigure}
    \begin{subfigure}{\textwidth}
        \includegraphics[width=\textwidth]{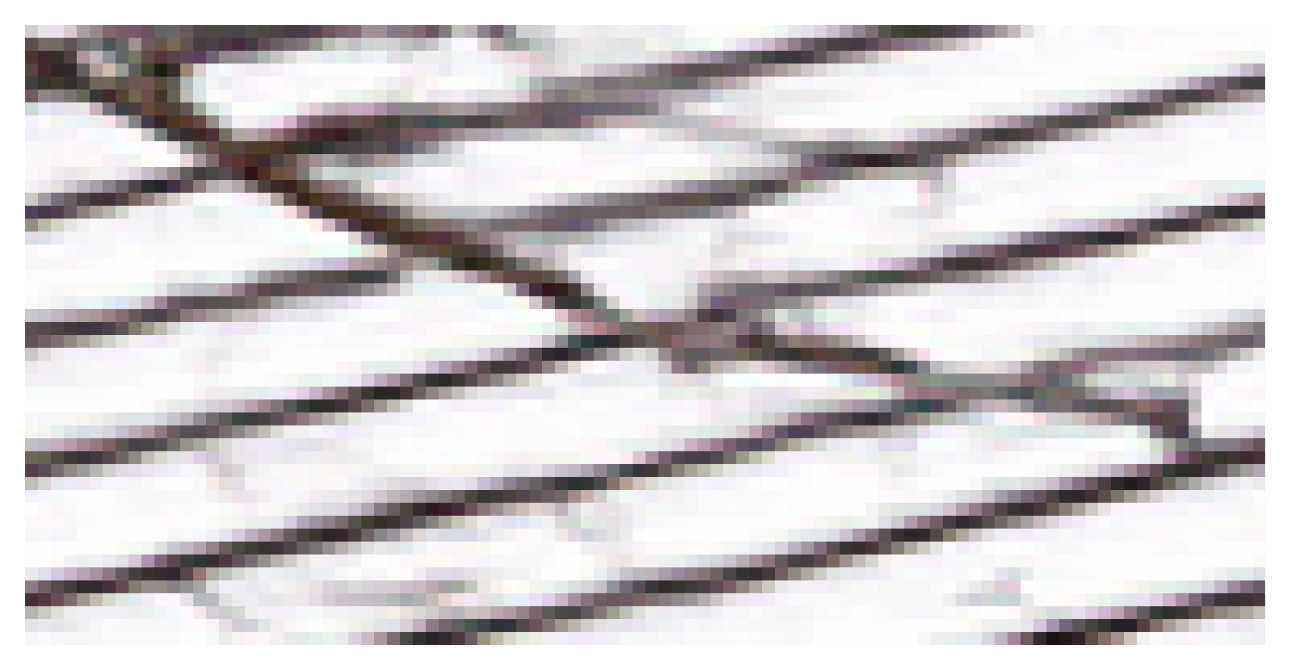}
    \end{subfigure}
    \begin{subfigure}{\textwidth}
        \includegraphics[width=\textwidth]{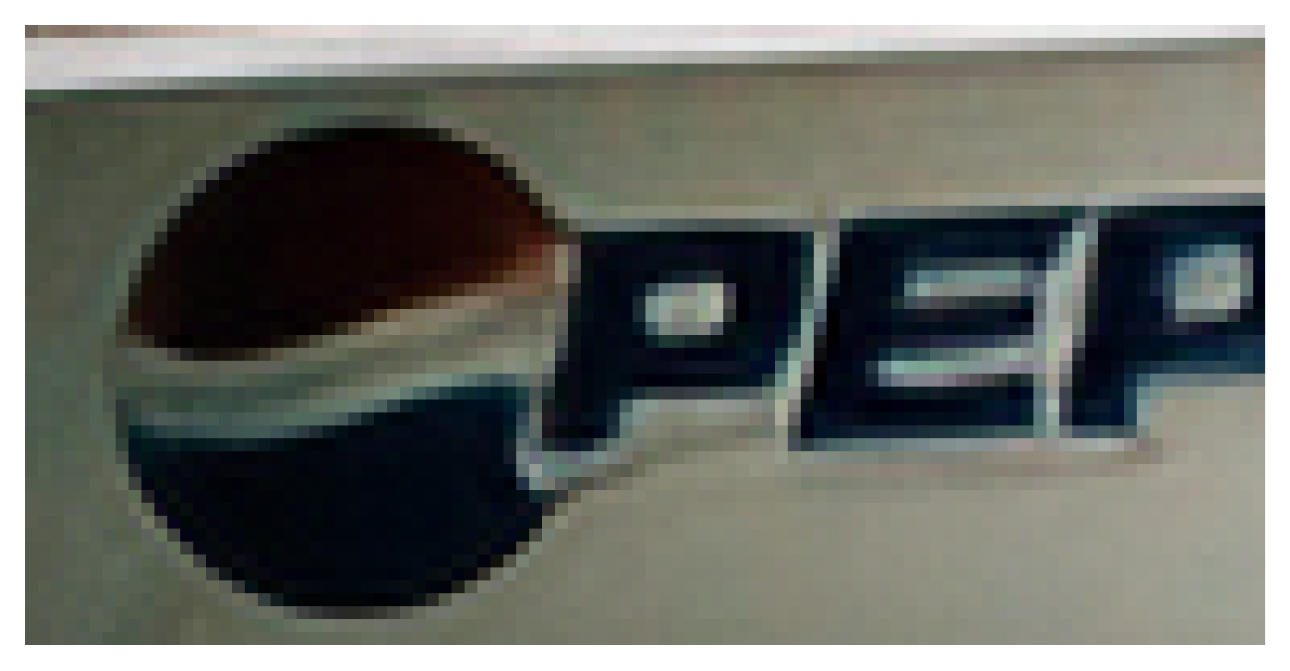}
    \end{subfigure} \vspace{-5pt}
\caption{Fusion-LPIPS}
\end{subfigure}
\begin{subfigure}{0.16\textwidth}
    \begin{subfigure}{\textwidth}
        \includegraphics[width=\textwidth]{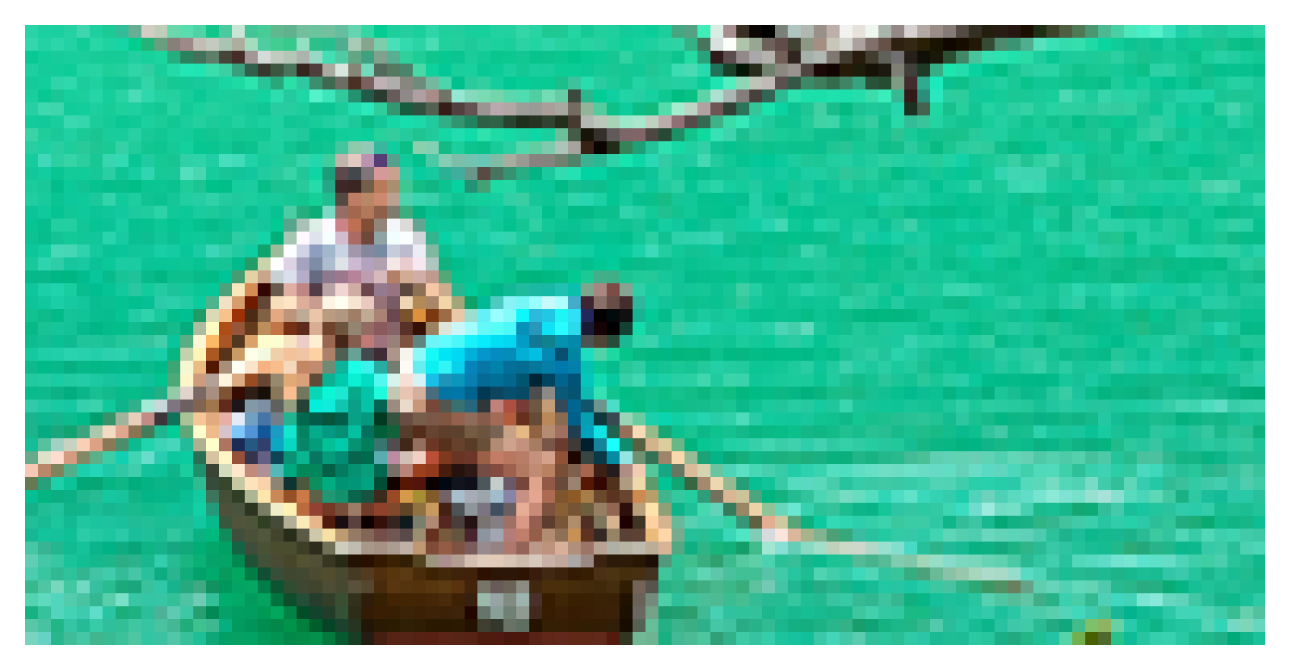}
    \end{subfigure}
    \begin{subfigure}{\textwidth}
        \includegraphics[width=\textwidth]{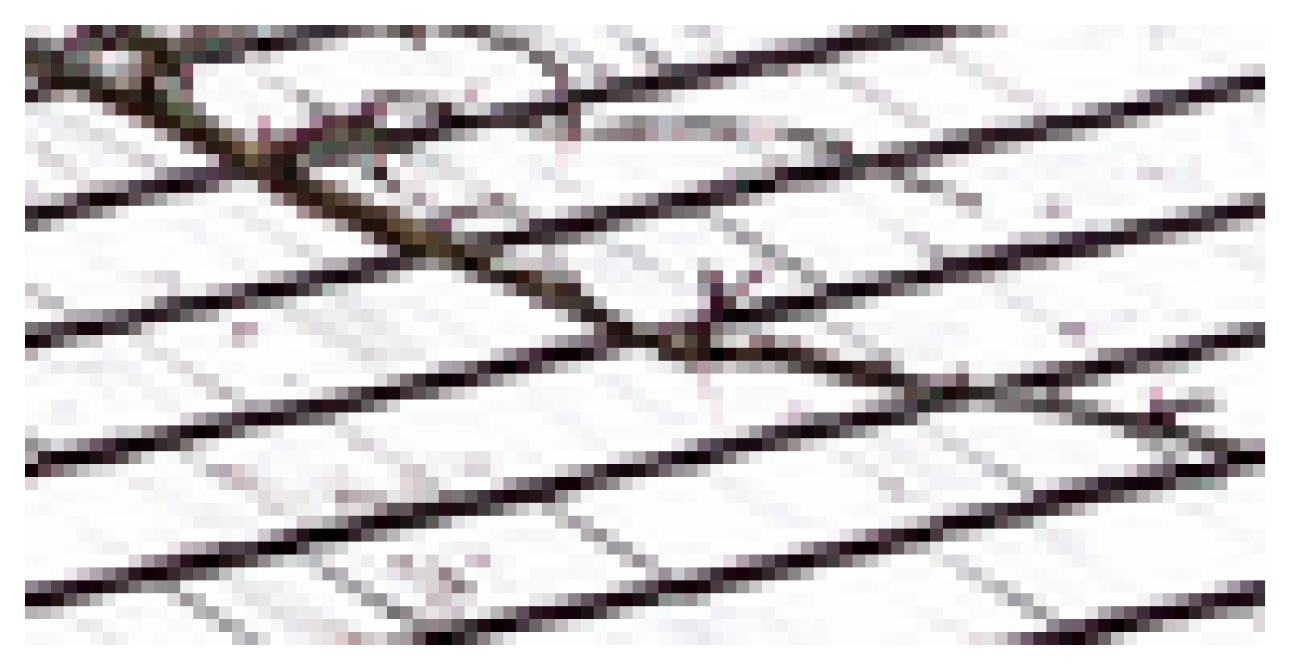}
    \end{subfigure}
    \begin{subfigure}{\textwidth}
        \includegraphics[width=\textwidth]{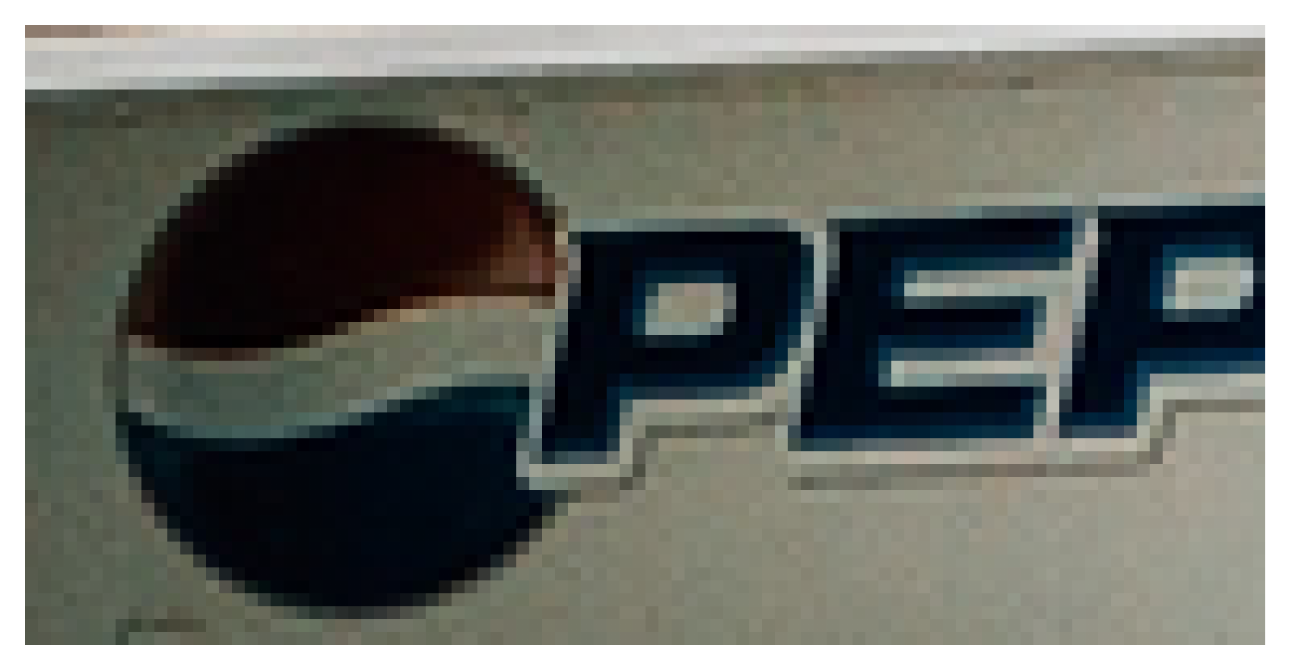}
    \end{subfigure} \vspace{-5pt}
\caption{Ground-truth HR}
\end{subfigure} \vspace{-8pt}
\caption{\textbf{Visual comparison of the proposed Fusion-LPIPS method with the state-of-the-art for $\times$4 SR on natural images from DIV2K validation set.} Fusion-LPIPS provides the best balance between fidelity and perceptual quality for natural images. }
\label{fig:natural_results} \vspace{-8pt}
\end{figure*}

\noindent \textbf{Perception-Distortion Trade-off.}
Perceptual loss \cite{10.1007/978-3-319-46475-6_43, wang2018esrgan} and adversarial loss \cite{NIPS2014_5ca3e9b1, wang2018esrgan, Zhang2019RankSRGANGA} have been introduced to remedy over-smoothing due to optimization of L1 or L2 loss. 
However, it has been shown that achieving high-fidelity and high perceptual quality (photo-realism) at the same time is infeasible, and there is a trade-off between distortion and perceptual quality of SR solutions~\cite{Blau_2018}. 
Adversarially trained GAN models offer a principled approach to obtain a trade-off between perception and distortion. However, they suffer from stability and convergence issues, and require fine-tuning of hyperparameters to get satisfactory results. Moreover, the GAN model is a deterministic mapping from \mbox{LR-to-HR image}; hence, a new model needs to be trained for each perception-distortion trade-off point. In contrast, we train a flow model only once, and propose methods to achieve multiple paths in the perception-distortion plane by simple post-processing of finitely many SR samples generated by a single model. \vspace{3pt}

\noindent \textbf{Flow-Based Generative Inference.} Normalizing flow models \cite{10.5555/3045118.3045281, Dinh2015NICENI, DBLP:conf/iclr/DinhSB17} exhibit several key advantages over GAN-based generative models, such as monotonic converge and stable training. SRFlow \cite{lugmayr2020srflow} and its extension SRFlow-DA \cite{jo2021srflowda} learn the distribution of plausible photo-realistic SR images conditioned on the LR image through log-likelihood training. They learn to transform a Gaussian distribution to the distribution of HR images conditioned on LR images, which can be formalized as: First learn an invertible mapping $\mathbf{z} = f(\mathbf{y};\mathbf{x})$, which maps an HR image $\mathbf{y}$ into latent variables $\mathbf{z}$, conditioned on the LR image $\mathbf{x}$. The probability density $p(\mathbf{y}|\mathbf{x})$ is computed explicitly by the log determinant of Jacobian of each invertible layer in the normalizing flow network \cite{Dinh2015NICENI, DBLP:conf/iclr/DinhSB17, rezende2016variational, kingma2018glow, lugmayr2020srflow}. To learn the parameters $\Theta$ of the network, the negative log-likelihood (NLL) loss is minimized. At inference time, photo-realistic images $\mathbf{y}$ can be generated by sampling from the latent encoding $\mathbf{z}$ as $\mathbf{y} = f^{-1}(\mathbf{z};\mathbf{x})$.
These models generate plausible, photo-realistic, but low fidelity images. In~this work, we use the normalizing flow model SRFlow-DA to generate sample SR solutions that define an SR space, and then propose a new strategy for navigating perception-distortion trade-off in this SR space to obtain high-fidelity photo-realistic solutions.

\vspace{-8pt}
\section{Perception-Distortion Trade-off by Image Fusion in the SR Space}
\label{sec:methodology}
\vspace{-4pt}

\noindent \textbf{Generating a Diverse Set of Samples in the SR Space. } \\
We employ the normalizing flow model SRFlow-DA \cite{jo2021srflowda} to generate a diverse set of SR images by sampling from the learned distribution at inference time. There are two parameters that need to be decided for the sampling process: (i) the temperature $\tau$ of the Gaussian density and (ii) the number of generated images. Infinitely many feasible SR images can be generated for each value of $\tau$ by sampling random seeds. If $\tau$ is close to zero, the generated SR sample images are all close to the mean of the distribution, hence the results achieve high fidelity scores but they appear blurry and lack fine details. In our experiments, we sample from a Gaussian distribution $\mathbf{z} \sim \mathcal{N}(0, \tau)$, with $\tau=0.9$, which yields a rich photo-realistic sample space. However, obtaining a single image in this SR space satisfying high-fidelity and/or photo-realism requirements of the application at hand is still a compelling problem. In the following, instead of searching for a single sample realization among infinitely many, we propose a novel image ensembling/fusion approach to achieve the best perception-distortion trade-off for the application at hand.\vspace{3pt}


\noindent \textbf{Fusion in the SR Space Favoring High Fidelity. }
SR image samples generated by flow models have rich texture, but suffer from poor fidelity. We propose a novel approach to combine a diverse set of SR samples into a single SR image either by ensembling them using simple pixelwise average or median operations over the samples in the ensemble dimension, or by fusing them using a simple ConvNet trained by L1 loss targeting high fidelity in the perception-distortion trade-off for information-centric applications. The~block diagram of the~proposed approach is depicted in Fig. \ref{fig:fusion_arch}. We average between 5-25 samples to control the trade-off path. The more samples averaged, the higher the fidelity. For the fusion network (Fusion-L1), we input a stack of $10$ SR samples. The output is a single RGB image. 
The average operation smooths out random variation in the samples due high-frequency hallucinated texture introduced by flow models and provides higher fidelity without compromising perceptual quality too much. Similar to averaging, computing the median also suppresses noise and achieves high fidelity SR outputs along with better perceptual results compared to individual flow samples.\vspace{3pt} 

\begin{figure*}[!t]
\centering
\begin{subfigure}{0.17\textwidth}
    \includegraphics[width=\textwidth]{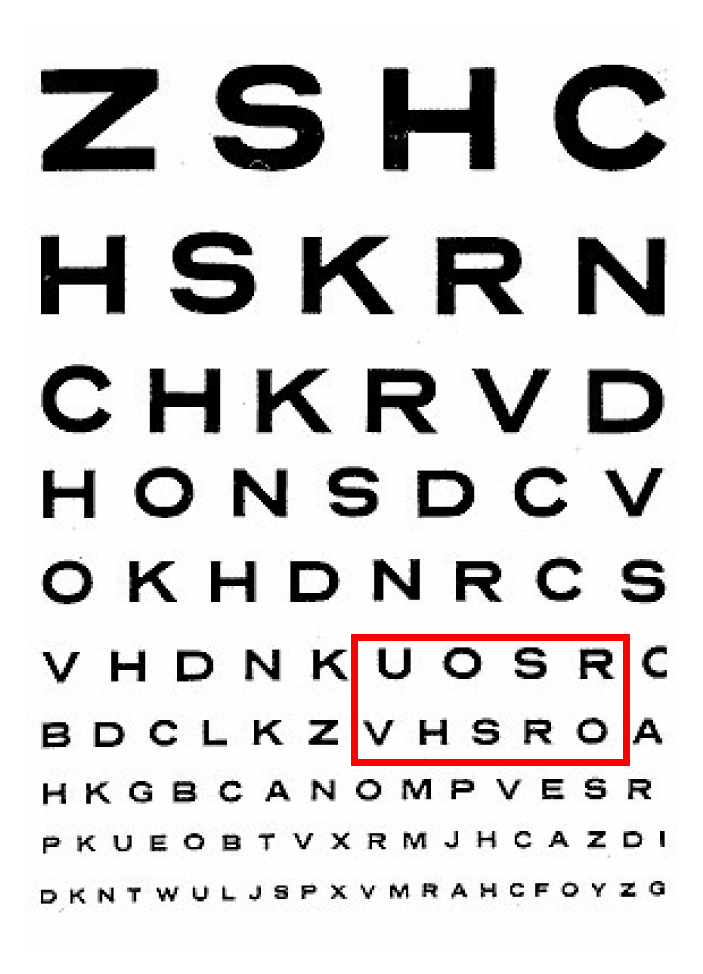} \vspace{-26pt} 
    \caption{Eye chart}
\end{subfigure}
\begin{subfigure}{0.19\textwidth}
    \begin{subfigure}{\textwidth}
        \includegraphics[width=\textwidth]{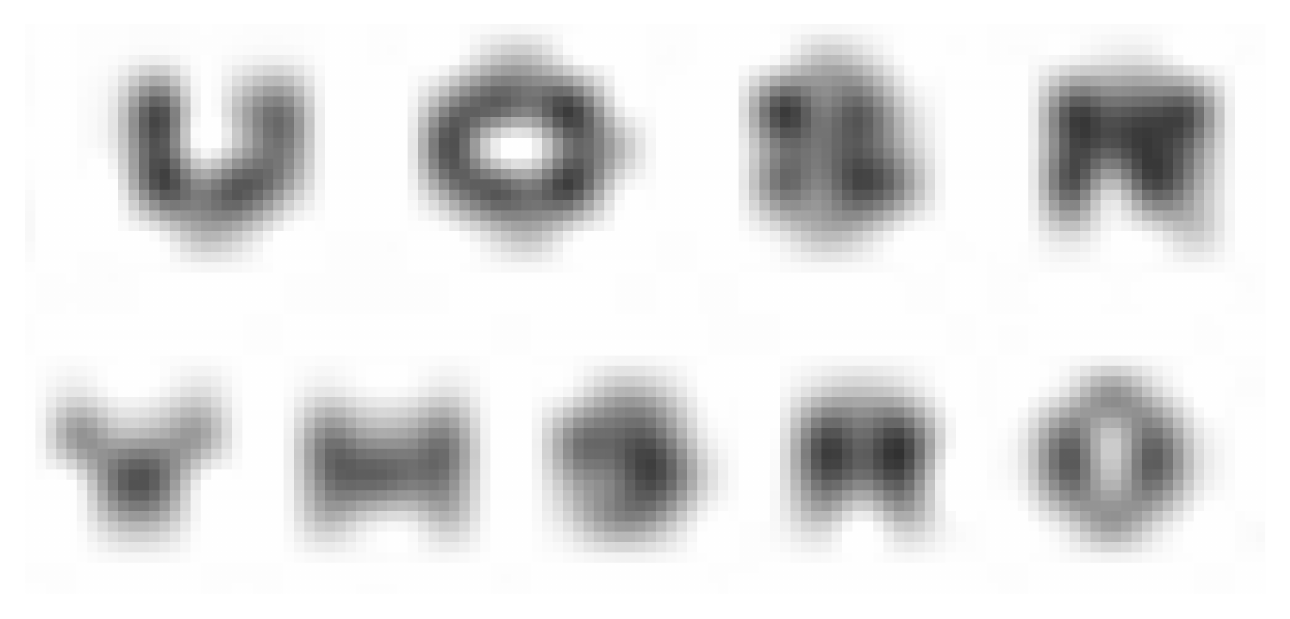}  \vspace{-14pt}
        \caption{LR}
    \end{subfigure}
    \begin{subfigure}{\textwidth}
        \includegraphics[width=\textwidth]{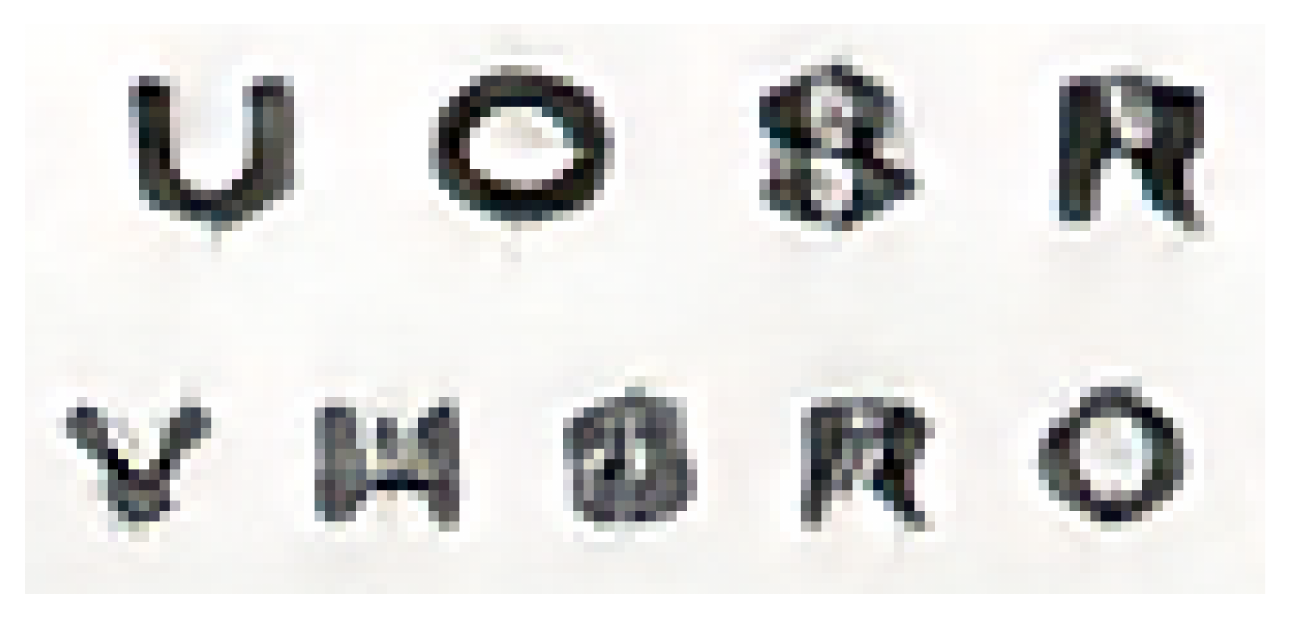}  \vspace{-14pt}
        \caption{ESRGAN+}
    \end{subfigure} 
\end{subfigure}
\begin{subfigure}{0.19\textwidth}
    \begin{subfigure}{\textwidth}
        \includegraphics[width=\textwidth]{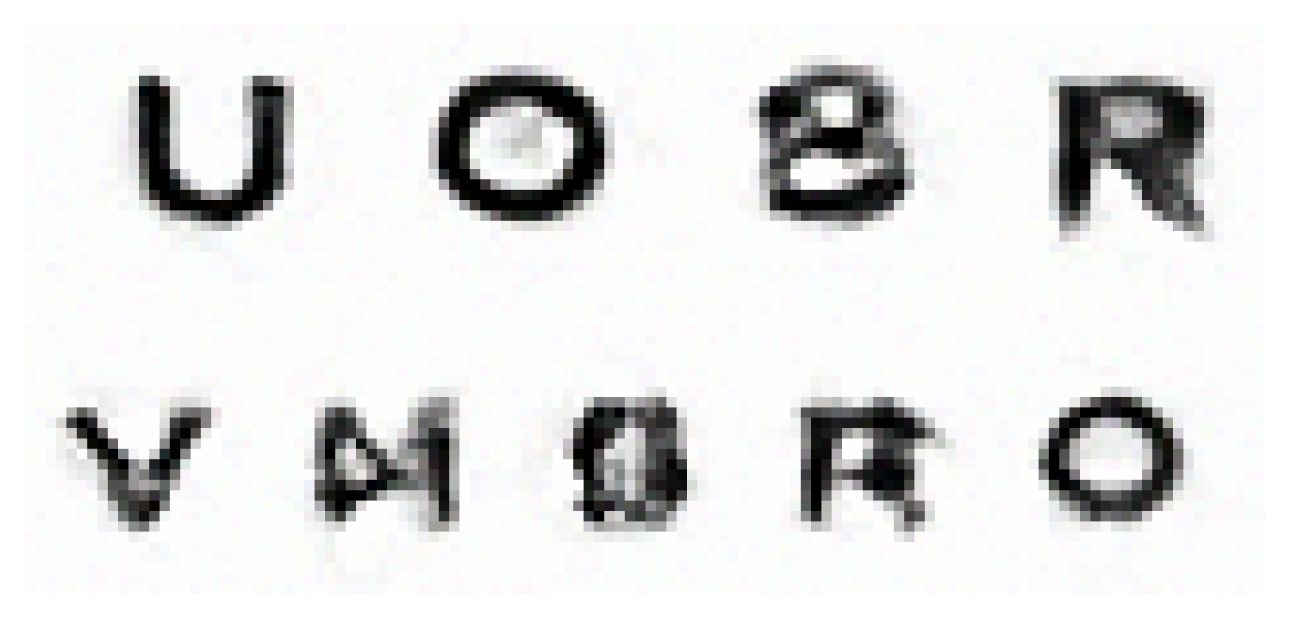}  \vspace{-14pt}
        \caption{Flow sample ($\tau$=0.9)}
    \end{subfigure}
    \begin{subfigure}{\textwidth}
        \includegraphics[width=\textwidth]{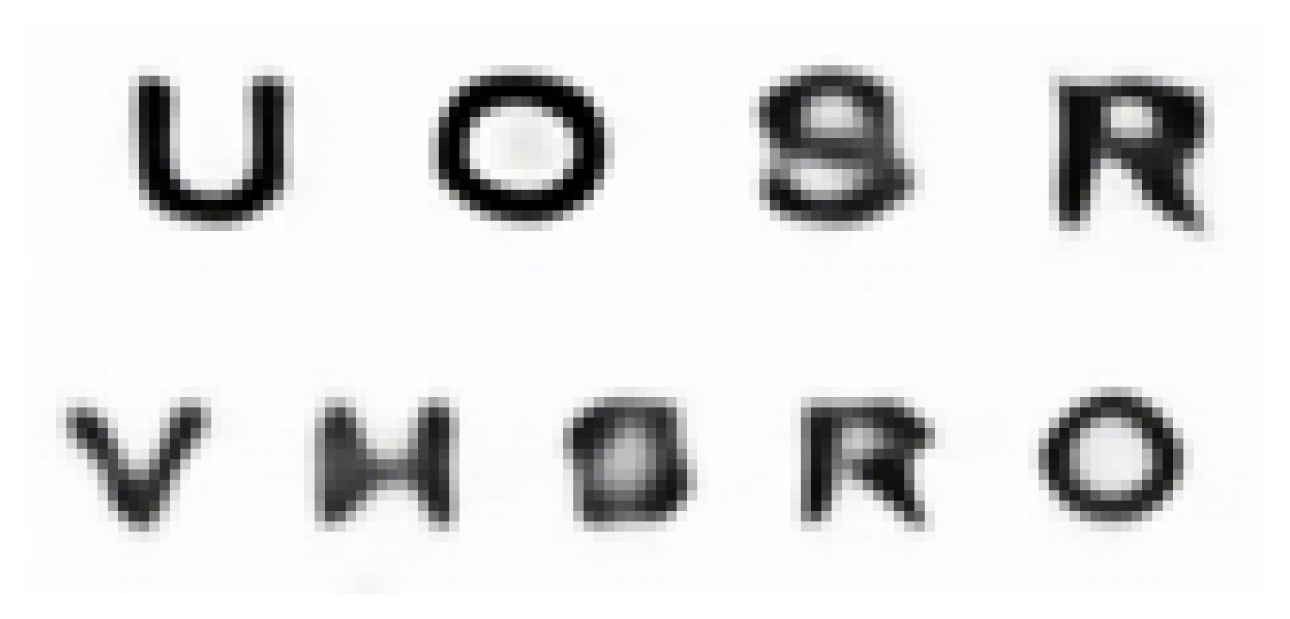}  \vspace{-14pt}
        \caption{Flow sample ($\tau$=0.1)}
    \end{subfigure}
\end{subfigure}
\begin{subfigure}{0.19\textwidth}
    \begin{subfigure}{\textwidth}
        \includegraphics[width=\textwidth]{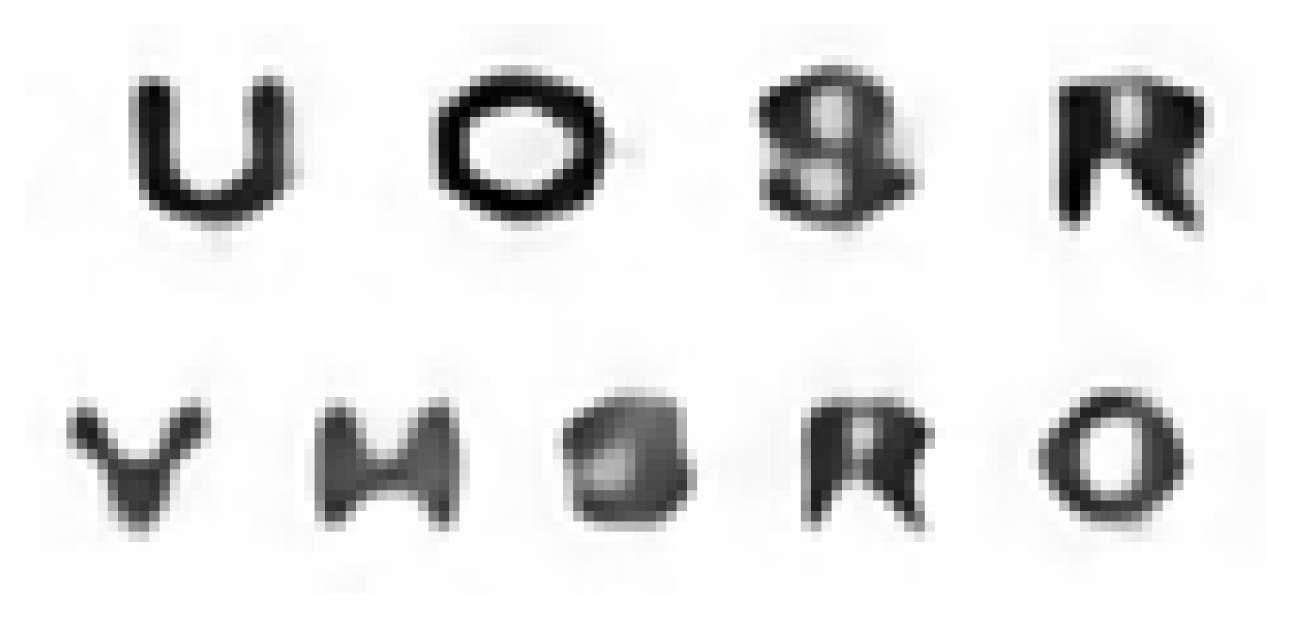} \vspace{-14pt}
        \caption{RRDB}
    \end{subfigure}
        \begin{subfigure}{\textwidth}
        \includegraphics[width=\textwidth]{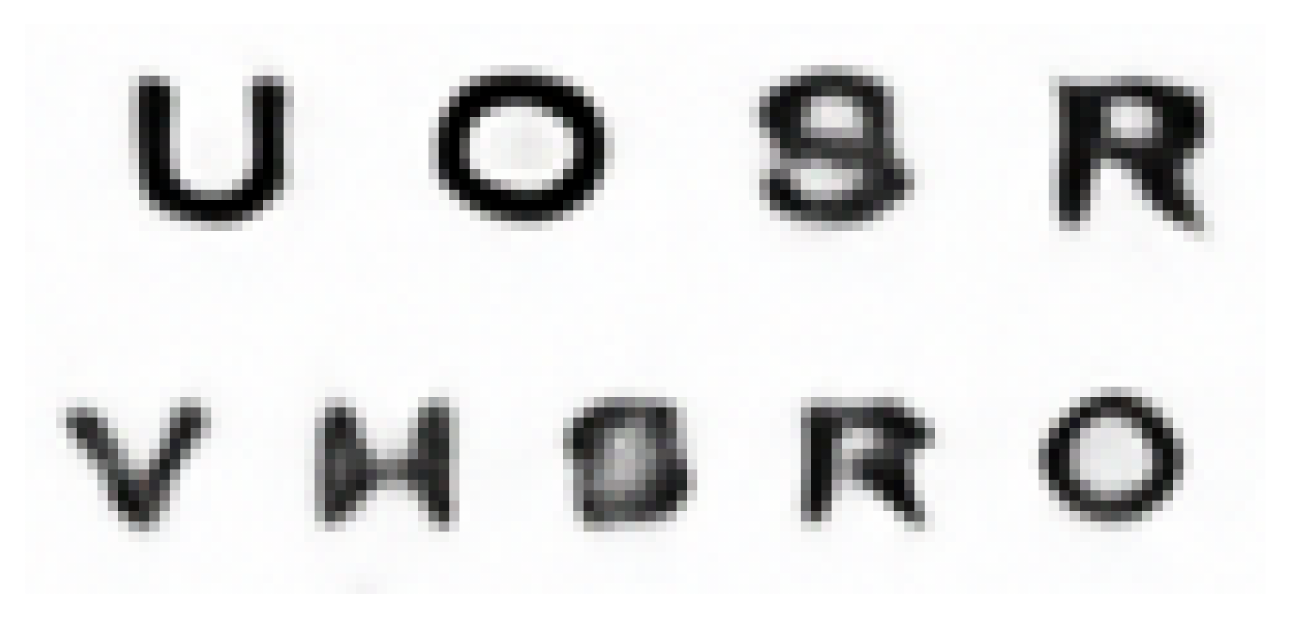} \vspace{-14pt}
       \caption{Average (25)}
    \end{subfigure}
\end{subfigure} 
\begin{subfigure}{0.19\textwidth}
    \begin{subfigure}{\textwidth}
        \includegraphics[width=\textwidth]{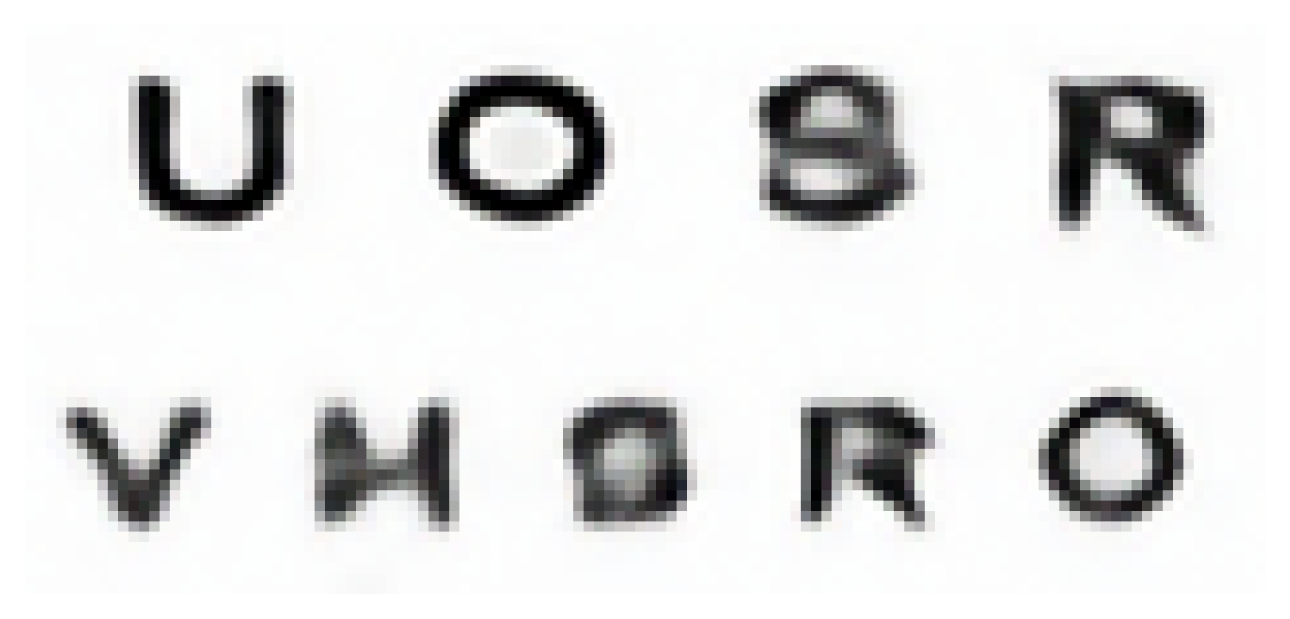} \vspace{-14pt}
        \caption{Fusion-L1}
    \end{subfigure}
    \begin{subfigure}{\textwidth}
        \includegraphics[width=\textwidth]{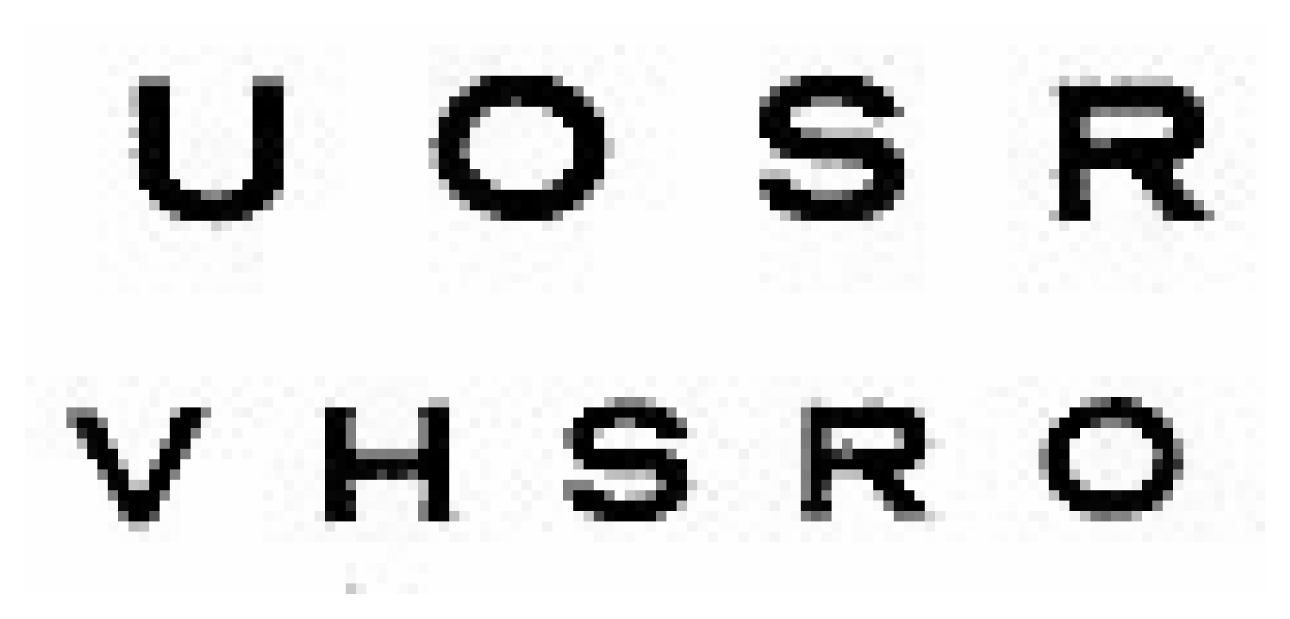} \vspace{-14pt}
        \caption{Ground-truth HR}
    \end{subfigure}
\end{subfigure} 
\vspace{-7pt}
\caption{\textbf{Visual comparison of the proposed Average and Fusion-L1 methods with the state-of-the-art for $\times$4 SR for a text image.} Average and Fusion-L1 methods favor high fidelity, which is desired in information-centric applications. }
\label{main_fig}
\end{figure*}

\begin{table*}[!t]
\begin{center}
\caption{\textbf{Performance comparison of $\times$4 SR models on DIV2K validation set}. The best and the second-best are marked in bold and underlined, respectively. Since SRFlow-DA generates multiple SR samples for a given LR image, SRFlow-DA scores are averages of the scores of multiple SR samples for each LR image in the validation set.
\vspace{-4pt}
}
\begin{adjustbox}{width=0.75\textwidth}
\begin{tabular}{lcccccccc}
\toprule
SR Model & PSNR$\uparrow$ & SSIM$\uparrow$ & LPIPS$_{\text{VGG}}$$\downarrow$ & PieAPP$\downarrow$ & DISTS$\downarrow$ & NIQE$\downarrow$ & MA$\uparrow$ & PI$\downarrow$ \\
\midrule
RRDB & 29.102 & \textbf{0.836} & 0.3050 & 1.1896 & 0.0480 & 4.768 & 4.507 & 4.869\\
ESRGAN+ & 26.001 & 0.725 & 0.2425 & 0.8485 & 0.0376 & 6.581 & 2.862 & \textbf{3.141} \vspace{0.075cm}\\

SRFlow-DA (0.1) & \textbf{29.269} & 0.824 & 0.2805 & 1.2074 & 0.0476 & 4.353 & 4.714 & 4.819 \\
SRFlow-DA (0.5) & 28.929 & 0.809 & 0.2531 & 1.1067 & 0.0432 & 4.789 & 5.326 & 4.732\\
SRFlow-DA (0.9) & 27.355 & 0.749 & 0.2534 & \underline{0.7820} & \underline{0.0354} & 4.071 & 5.079 & 4.496\vspace{0.075cm}\\
Average (5) & 28.882 & 0.809 & \underline{0.2378} & 0.9990 & 0.0396 & \underline{3.906} & 5.614 & 4.146 \\
Average (25) & \textbf{29.269} & \underline{0.825} & 0.2615 & 1.1002 & 0.0441 & 4.049 & 5.025 & 4.512 \\
Median (5) & 28.721 & 0.804 & 0.2399 & 0.9969 & 0.0393 & 4.221 & \underline{5.747} & 4.237 \\
Median (25) & 29.222 & 0.823 & 0.2564 & 1.1085 & 0.0443 & 4.071 & 5.079 & 4.496 \\
Fusion-L1 & \underline{29.249} & 0.823 & 0.2833 & 1.1384 & 0.0478 & 4.524 & 4.914 & 5.195 \\
Fusion-LPIPS & 28.281 & 0.794 & \textbf{0.2151} & \textbf{0.7301} & \textbf{0.0337} & \textbf{3.270} & \textbf{6.416} & \underline{3.427}\\
\bottomrule
\end{tabular}
\end{adjustbox}
\label{table:fusion_results}
\end{center}  
\vspace{-10pt}
\end{table*}

\noindent \textbf{Fusion in the SR Space Favoring Perceptual Quality. }
When the target test set contains natural images with significant texture (as opposed to text or numerals), and the~evaluation criterion is perceptual quality (rather than information recovery), the perception-distortion tradeoff can be steered towards the perception dimension by training the fusion network using LPIPS loss~\cite{zhang2018perceptual}, resulting in the Fusion-LPIPS model. LPIPS loss can be expressed as the~distance between two patches $\mathbf{x}$ and $\mathbf{x}_0$ passing through a pretrained network $F$:
%
\[ d(\mathbf{x}, \mathbf{x}_0) = \sum_l \frac{1}{H_{l}W_{l}} \sum_{h,w} || w_l \odot (\hat{y}_{hw}^{l} - \hat{y}_{0hw}^{l}) ||_2^{2}, \]
\noindent
where activations of a selected layer of network $F$ are normalized along the channel dimension $l$, and then scaled by vector~$w_l$. LPIPS loss enforces preserving feature-level similarities and accounts for many nuances of human perception; thus, the Fusion-LPIPS model provides natural-looking images by restoring realistic textures accurately.
Our results show that the proposed Fusion-LPIPS network yields results that are perceptually similar or superior to the results that can be obtained by the state-of-the-art adversarial approaches while providing a significant improvement in fidelity. 


\section{Experimental Results} \vspace{-4pt}
\noindent \textbf{Implementation Details.} 
In our experiments, we use the~standard bicubic LR model settings so that a total of 3450 training LR images are generated from DIV2K \cite{Agustsson_2017_CVPR_Workshops} and Flickr2K \cite{EDSR2017} datasets by using the MATLAB bicubic downsampling kernel. HR~patches of 160$\times$160 pixels are cropped and Gaussian noise with $\sigma = {4}/{\sqrt{3}}$ is added to HR patches for $\times$4 SR.
In order to train the generative model SRFlow-DA, we follow the same procedure described in~\cite{jo2021srflowda}.  As~the~low-resolution encoding network, the pre-trained RRDB model~\cite{wang2018esrgan} with 23 blocks is selected to capture the underlying representation of the LR image. Once the SRFlow-DA model is trained, we sample from a Gaussian distribution, specifically $\mathbf{z} \sim \mathcal{N}(0, \tau)$, where $\tau$ is set to 0.9.
 
The Fusion-L1 and Fusion-LPIPS network architectures consist of two residual blocks, where the number of output channels is set to 64. We train the Fusion-L1 network using L1 loss and the Fusion-LPIPS network using LPIPS loss for 10K iterations. We~use the Adam  optimizer~\cite{DBLP:conf/iclr/DinhSB17} with an initial learning rate of $2 \times 10^{-4}$ and a cosine annealing scheduler to adjust the~learning rate during training.\vspace{3pt}

\noindent \textbf{Comparison with the State-of-the-Art.} We compare our image ensembling/fusion strategy with RRDB \cite{wang2018esrgan} (high fidelity), ESRGAN+ \cite{esrganplus} (high perceptual quality),
and samples of SRFlow-DA \cite{jo2021srflowda} with different values of $\tau$ to evaluate perception-distortion performance on the DIV2K validation set \cite{Agustsson_2017_CVPR_Workshops}. All compared methods are also trained on DF2K dataset~\cite{EDSR2017}, a combination of DIV2K and Flickr2K, by using publicly available codes. 

Quantitative results are shown in Table~\ref{table:fusion_results}. PSNR and SSIM are used to evaluate fidelity, while LPIPS, PieAPP, DISTS, NIQE, Ma, and PI scores are used to measure perceptual quality (photo-realism). Our proposed image ensembling by averaging obtains the best PSNR among all solutions. As the number of samples averaged varies from 5 to 25, the fidelity measures improve, while perceptual measures slightly degrade, providing a controllable perception-distortion tradeoff path.
On the other hand, our proposed Fusion-LPIPS model consistently outperforms flow model output samples in all perceptual metrics while also providing higher fidelity. 

Qualitative results depicted in Fig.~\ref{fig:natural_results} and Fig.~\ref{main_fig} show that our proposed solutions provide the highest visual quality SR images. As expected, high PSNR values for RRDB do not correspond to high perceptual quality. On the other hand, ESRGAN+ and SRFlow-DA frequently produce visible artifacts and excessive sharpness. Compared to sample SRFlow-DA images, the Fusion-LPIPS model avoids hallucination of high-frequency details, thus successfully generates photo-realistic images. As~shown in Fig.~\ref{main_fig}, when aiming for high fidelity, pixelwise averaging or the fusion model trained with L1 loss gives more satisfactory results as compared to the existing SR~models.
\vspace{-10pt}
\section{Conclusion} \vspace{-6pt}
We propose a simple but effective image ensembling/fusion approach to obtain a single SR image with the desired perception-distortion trade-off benefiting from a diverse set of feasible photo-realistic solutions in the SR space spanned by flow models. Different image ensembling and fusion strategies are proposed offering multiple paths to move sample solutions in the SR space to more desired positions in the perception-distortion plane depending on the fidelity vs. perceptual quality requirements of the task at hand. 

Our approach achieves promising results on the DIV2K validation set in terms of both quantitative metrics and visual quality. Averaging or median of samples pixelwise in the ensemble dimension gives the best PSNR and the second best MS-SSIM score without significantly compromising perceptual quality, with PI score very close to that of SRFlow $\tau =0.9$ samples. The Fusion-LPIPS model is able to achieve improved results compared to state-of-the-art adversarial approaches.  Indeed, it achieves the second best result PI measure behind ESRGAN+, with 2.28 dB improvement in PSNR over the ESRGAN+ model. The results indicate that our proposed SR sample fusion strategy has the potential for finding solutions that have better perception-distortion trade-off.
 





\clearpage

\bibliographystyle{IEEEbib}
\bibliography{source}

\end{document}